\documentclass[sn-mathphys,Numbered]{sn-jnl}%

\usepackage{graphicx}%
\usepackage{multirow}%
\usepackage{amsmath,amssymb,amsfonts}%
\usepackage{amsthm}%
\usepackage{mathrsfs}%
\usepackage[title]{appendix}%
\usepackage{xcolor}%
\usepackage{textcomp}%
\usepackage{manyfoot}%
\usepackage{booktabs}%
\usepackage{algorithm}%
\usepackage{algorithmic}
\usepackage{listings}%
\usepackage{tabularx}

\usepackage{subfigure}

\usepackage{setspace}

\usepackage{wrapfig}
\usepackage{algorithm}
\usepackage{listings}
\usepackage{color}

\newcommand{\vx}{\mathbf{x}}
\newcommand{\vy}{\mathbf{y}}
\newcommand{\name}{\textsc{COURIER}}

\newcommand{\refeq}[1]{Eq. (\ref{#1})}
\newcommand{\reffig}[1]{Fig. \ref{#1}}

\newcommand{\reftab}[1]{Table. (\ref{#1})}
\newcommand{\refsec}[1]{Section. \ref{#1}}

\begin{document}

\title[COURIER: Contrastive User Intention Reconstruction]{COURIER: Contrastive User Intention Reconstruction for Large-Scale Visual Recommendation}

\author[1,2]{\fnm{Jia-Qi} \sur{Yang}}\email{yangjq@lamda.nju.edu.cn}
\author[3]{\fnm{Chenglei} \sur{Dai}}\email{chenglei.dcl@taobao.com}
\author[3]{\fnm{Dan} \sur{OU}}\email{oudan.od@taobao.com}
\author[3]{\fnm{Dongshuai} \sur{Li}}\email{lidongshuai.lds@taobao.com}
\author[3]{\fnm{Ju} \sur{Huang}}\email{huangju.hj@taobao.com}
\author*[1,2]{\fnm{De-Chuan} \sur{Zhan}}\email{zhandc@nju.edu.cn}
\author[3]{\fnm{Xiaoyi} \sur{Zeng}}\email{yuanhan@taobao.com}
\author*[4]{\fnm{Yang} \sur{Yang}}\email{yyang@njust.edu.cn}

\affil[1]{\orgname{School of Artificial Intelligence, Nanjing University}, \city{Nanjing 210023}, \country{China}}
\affil[2]{\orgname{State Key Laboratory for Novel Software Technology, Nanjing University}, \city{Nanjing 210023}, \country{China}}
\affil[3]{\orgname{Alibaba Group}, \city{Hangzhou}, \country{China}}
\affil[4]{\orgname{Nanjing University of Science and Technology}, \city{Nanjing}, \country{China}}

\abstract{
    With the advancement of multimedia internet, the impact of visual characteristics on the decision of users to click or not within the online retail industry is increasingly significant.
    Thus, incorporating visual features is a promising direction for further performance improvements in click-through rate (CTR). However, experiments on our production system revealed that simply injecting the image embeddings trained with established pre-training methods only has marginal improvements.
    We believe that the main advantage of existing image feature pre-training methods lies in their effectiveness for cross-modal predictions. However, this differs significantly from the task of CTR prediction in recommendation systems. In recommendation systems, other modalities of information (such as text) can be directly used as features in downstream models. Even if the performance of cross-modal prediction tasks is excellent, it is challenging to provide significant information gain for the downstream models.
    We argue that a visual feature pre-training method tailored for recommendation is necessary for further improvements beyond existing modality features.
    To this end, we propose an effective user intention reconstruction module to mine visual features related to user interests from behavior histories, which constructs a many-to-one correspondence.
    We further propose a contrastive training method to learn the user intentions and prevent the collapse of embedding vectors.
    We conduct extensive experimental evaluations on public datasets and our production system to verify that our method can learn users' visual interests. Our method achieves $0.46\%$ improvement in offline AUC and $0.88\%$ improvement in Taobao GMV (Cross Merchandise Volume) with p-value$<$0.01.}

\keywords{User Intention Reconstruction, Contrastive Learning, Personalized Searching, Image Features}

\maketitle

\section{Introduction}

Predicting the click-through rate (CTR) is an essential task
in recommendation systems based on deep learning\cite{ctr_a,ctr_b_dien,gc_calib}.
Typically, CTR models measure the probability that a user will click on an item
based on the user's profile and behavior history\cite{ctr_b_dien,gdfm},
such as clicking, purchasing, adding a cart, etc.
The behavior histories are represented by sequences of
item IDs, titles, and some statistical features, such as monthly sales
and favorable rate\cite{ctr_a}.

From an intuitive perspective, the visual representations of items hold significant importance in online item recommendation, particularly in categories like women's clothing.
Recent studies have shown that incorporating modality information through end-to-end training can enhance the effectiveness of recommendations\cite{e2e_rec_iclr23}. However, when dealing with large-scale product recommendation systems, where users generate billions of clicks daily, implementing end-to-end training becomes impractical\cite{zhou2023-throughput}.
A feasible approach involves enhancing image features through the utilization of representation learning methods specifically designed for image features. Nevertheless, our experiments reveal that employing embeddings derived from existing image feature learning methods such as SimCLR\cite{simclr}, SimSiam\cite{simsiam}, and CLIP\cite{clip}, only yield marginal effects on the relevant downstream CTR prediction task (\reftab{tab:main}).
We ascribe this lack of success to two factors. Firstly, in the context of recommendation scenarios, user preferences for visual appearance tend to be vague and imprecise, which may not be captured by existing image feature learning methods that focus on label prediction.
Besides, augmentation methods that excel in tasks like image classification, due to their significant alteration of the image's appearance, are not suitable for recommendation systems.
Secondly,
the label information embedded in the pre-trained embeddings, such as classes or language descriptions, can already be directly utilized in item recommendations, rendering the information gain provided by pre-trained embeddings redundant. For instance, we already possess categories, item titles, and style tags provided by the merchants in Taobao.
Consequently, a pre-trained model that performs well in predicting categories or titles contributes little novel information and does not enhance CTR prediction task.

\begin{figure}
    \centering
    \includegraphics[width=\linewidth]{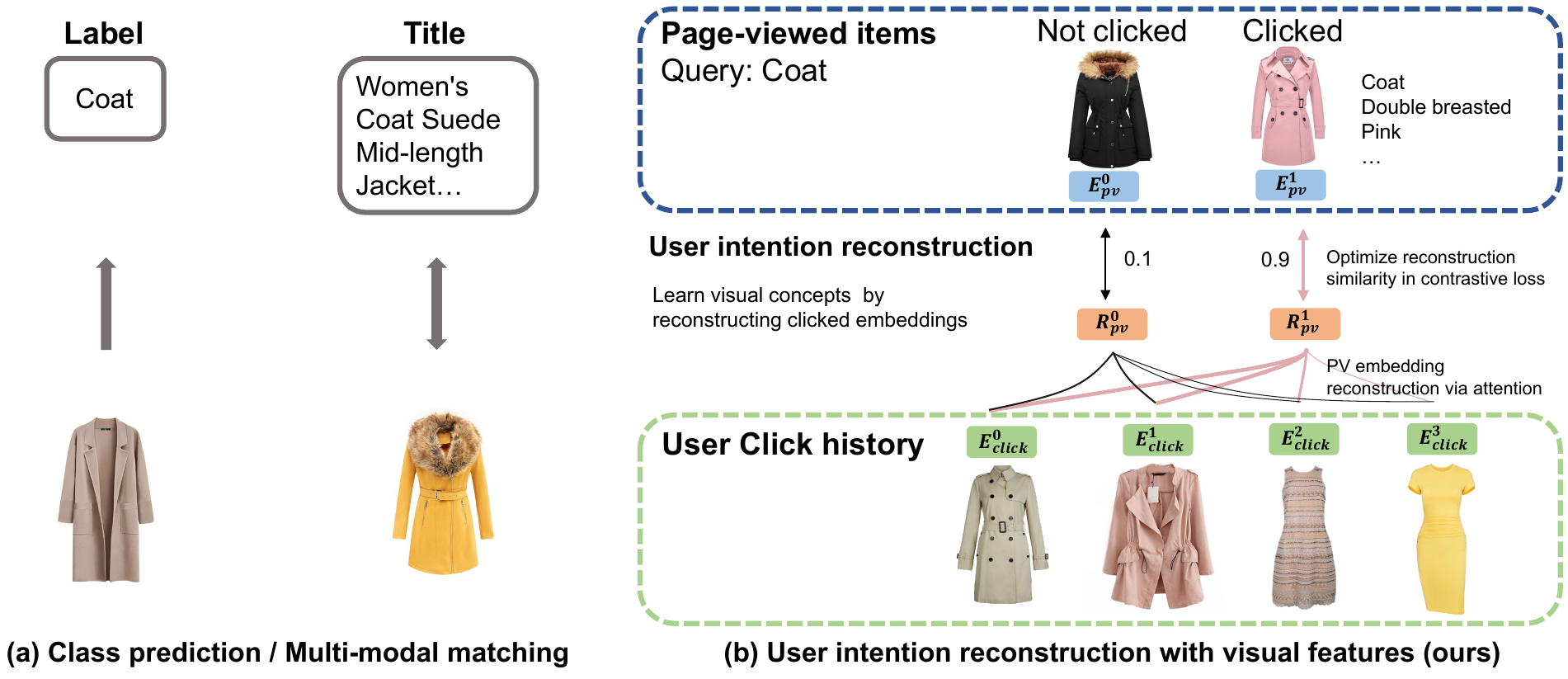}
    \caption{
        (a) Existing image feature learning methods are tailored for cross-modal prediction tasks.
        (b) We propose a user intention reconstruction method to mine potential visual features that cannot be reflected by cross-modal labels. 
        In this example, the user searched for "Coat" and received two recommendations (Page-viewed items). 
        The user clicked on the one on the right. 
        Through our user intention reconstruction, we identified similar items from the user's click history with larger attention, the reconstructed PV item embeddings are 
        denoted as $R_{pv}^j$. Then, we optimize the PV embeddings $E_{pv}^j$ and reconstructions $R_{pv}^j$ to be closer if the corresponding item is clicked and more far apart otherwise.
        \label{fig:intro}
    }
\end{figure}

To boost the performance of downstream CTR prediction tasks, we argue that the representation learning method should be aware of the downstream CTR task and
should also be decoupled from the CTR task to reduce computation.
To achieve this goal,
we propose a \textbf{CO}ntrastive \textbf{U}se\textbf{R} \textbf{I}nt\textbf{E}ntion
\textbf{R}econstruction (\textbf{\name{}}) method.
Our method is based on an intuitive assumption:
An item clicked by a user is likely to have
visual characteristics similar to some of the clicking history items.
One straightforward approach is to optimize the distance between user embeddings (comprised of item images previously clicked by the user) and clicked item image embeddings. 
Unlike the typical one-to-one correspondence in common contrastive learning, this establishes a many-to-one correspondence. Consequently, we must aggregate multiple item image embeddings from the user's historical clicks. Common aggregation methods include self-attention or pooling.
However, it should be noted that user click histories often contain a significant number of images that have low relevance to the clicked items. Directly minimizing the distance between these aggregated embeddings may result in all images having very similar embeddings.
To mine the visual features related to user interests,
we propose reconstructing the next clicking item with
a cross-attention mechanism on the clicking history items.
The reconstruction can be interpreted by
a weighted sum of history item embeddings, which effectively selects related images from history in an end-to-end manner,
as depicted in \reffig{fig:intro}.
We propose to optimize a contrastive loss
that not only encourages lower reconstruction error,
but also push embeddings of un-clicked items further apart.
We conducted various experiments to verify our motivation and
design of the method.
Our pre-trained image embedding achieves $12\%$ improvements in NDCG and Recall on several public datasets compared with strong baselines.
We conducted experiments in a large-scale Taobao dataset, our method achieves $0.46\%$ absolute offline improvement on AUC.
In online A/B tests, we achieve $0.88\%$ improvement
on GMV in the women's clothing category,
which is significant considering the volume of Taobao online shopping platform.

Our contribution can be summarized as follows:
\begin{itemize}
    \item To establish a one-to-one correspondence between images clicked by the user in the past and the image currently clicked by the user, we propose a user intention reconstruction method, which can mine latent user intention from history click sequences without any explicit semantic labels.
    \item The user intention reconstruction objective alone may lead to a collapsed solution. To solve this problem, we propose a contrastive training method that utilizes the un-clicked item efficiently.
    \item We conduct comprehensive experiments on both private and public datasets to validate the effectiveness of our method. Additionally, we provide insights and share our practical experience regarding the deployment of image feature models in real-world large-scale recommendation systems.
\end{itemize}

\section{Related work}

From collaborative filtering\cite{cf_rec,amazon} to deep learning based
recommender systems\cite{deep_rec,dssm,esdfm},
IDs and categories (user IDs, item IDs, advertisement IDs, tags, etc.) are
the essential features in item recommendation systems,
which can straightforwardly represent the identities.
However, with the development of the multi-media internet,
ID features alone can hardly cover all the important
information.
Thus, recommendations based on content such as
images, videos, and texts have become an active
research field in recommender systems.
In this section, we briefly review the related work
and discuss their differences compared with our method.

\subsection{Content-based recommendation}

In general, content is more important when the content
itself is the concerned information.
Thus, the content-based recommendation has already been
applied in news\cite{news_a},
music\cite{music_rec},
image\cite{DBLP:journals/tkde/WuCHFXW20},
and video\cite{youtube} recommendations.

Research on the content-based recommendation in the item recommendation task
is much fewer because of the dominating ID features.
Early applications of image features typically use
image embeddings extracted from a pre-trained image classification model,
such as \cite{no_e2e_image_rec,mmgcn,dualgnn,lattice,slmrec,mgcn,bm3}.
The image features adopted by these methods are trained on
general classification datasets such as ImageNet,
which may not fit recommendation tasks.
Thus, \cite{M5Product} proposes to train on multi-modal data
from recommendation task.
However, \cite{M5Product} does not utilize any recommendation labels
such as clicks and payments,
which will have marginal improvement if we are already using information from
other modalities.

With the increase in computing power,
some recent papers propose to train item recommendation models end-to-end
with image feature networks\cite{e2e_rec_iclr23,multi_modal_rec}.
However, the datasets used in these papers are much smaller than our application scenario.
For example, \cite{multi_modal_rec} uses a dataset consisting of 25 million interactions.
While in our online system, average daily interactions in a single category (women's clothing) are
about 850 million.
We found it infeasible to train image networks in our scenario end-to-end,
which motivates our decoupled two-stage framework with user-interest-aware embedding learning.

\subsection{Image representation learning in recommendation}

Self-supervised pre-training has been a hot topic in recent years.
We classify the self-supervised learning methods into two categories:
Augmentation-based and prediction-based.

\noindent \textbf{Augmentation-based}.
The augmentation-based methods generate multiple different views of
an image by random transformations,
then the model is trained to pull the embeddings of different views
closer and push other embeddings (augmented from different images) further.
SimCLR\cite{simclr}, SimSiam\cite{simsiam}, BYOL\cite{byol} are
some famous self-supervised methods in this category.
These augmentation-based methods do not perform well in the item recommendation task
as shown in our experiments in \refsec{sec:perform_ctr}.
Since the augmentations are designed for classification (or segmentation, etc.) tasks,
they change the visual appearance of the images without changing their
semantic class,
which contradicts the fact that visual appearance is also important in recommendations (e.g., color, shape, etc.).

\noindent \textbf{Prediction-based}.
If the data can be split into more than one part,
we can train a model taking some of the parts as inputs and predict
the rest parts,
which is the basic idea of prediction-based pre-training.
Representative prediction-based methods include
BERT\cite{bert}, CLIP\cite{clip}, etc.
The prediction-based methods can be used to train multi-modal recommendation
data as proposed by \cite{M5Product}.
However, if we are already utilizing multi-modal information,
the improvements are limited, as shown in our experiments.
To learn user interests information that can not be provided by
other modalities,
we argue that user behaviors should be utilized, as in our proposed method.

\subsection{Contrastive learning in recommender systems}

Contrastive learning methods are also adopted in recommendation systems
in recent years.
The most explored augmentation-based method is augmenting data by dropping,
reordering, and masking some features\cite{contras_cikm21}, items\cite{contras_wsdm22,contras_kdd22_x}, and graph edges\cite{contras_sigir21}.
Prediction-based methods are also adopted for recommendation tasks,
e.g., BERT4Rec\cite{bert4rec} randomly masks some items and makes predictions.
However, all these recommender contrastive learning methods
concentrate on augmentation and pre-training with
ID features, while our method tackles representation learning of
image features.
Several recent works also considered mining users' intents with contrastive learning\cite{intent_ijcai,intent_sigir}.
Different from our concentration on visual features, 
they focus on learning with graph structures while image features are not considered.

\section{Contrastive user intention reconstruction}

We briefly introduce some essential concepts and notations,
then introduce our method in detail.

\subsection{Preliminary}

\noindent \textbf{Notations}.
A data sample for CTR prediction in item search
can be represented with a tuple of (user, item, query, label).
Typically, in recommendation tasks, there is no explicit query, and the remaining aspects align with search. Our approach is universally applicable in these scenarios. For simplicity, the subsequent sections will consistently use the term ``recommendation''.
A user searches for a query text,
and several items are shown to the user.
Then the items that the user clicked are labeled as positive
and negative otherwise.
When a user views a page,
we have a list of items that are presented to the user,
we call them page-view (PV) items.
The length of this PV list is denoted by $l_{pv}$.
Each PV item has a cover image, denoted by $I_{pv}^j$,
where $0 \le j < l_{pv}$.
The corresponding click labels are denoted by $y_{pv}^j$, where $y_{pv}^j\in \{0, 1\}$.
Each user has a list of clicked item history,
the image of each item is denoted by $I_{click}^k$.
$0 \le k < l_{click}$, where $l_{click}$ is the length of
click history.
The $l_{pv}$ and $l_{click}$ may vary by different users and pages,
we trim or pad to the same length practically.

\noindent \textbf{Attention}.
An attention layer is defined as follows:
\begin{equation}
    \operatorname{Attention}(Q, K, V) = \operatorname{softmax}(\frac{Q K^T}{\sqrt{d_K}})V
\end{equation}
where $Q$ %
is the query matrix,
$K$ %
is the key matrix,
$V$ %
is the value matrix.
The mechanism of the attention layer can be interpreted
intuitively:
For each query, a similarity score is computed with
every key,
the values corresponding to keys are weighted by their
similarity scores and summed up to obtain the outputs.
We refer interested readers to \cite{attn} for
more details of the attention mechanism.

\subsection{User intention reconstruction\label{sec:reconstruction}}

In the following discussion,
we only consider a single line of data (a PV list and corresponding list of user click history),
the batch training method will be discussed later.
We use $I_{pv}$ and $I_{click}$ to denote the matrix of
all the $l_{pv}$ and $l_{click}$ images.
All the images are fed to the image backbone (IB) to
get their embeddings.
We denote the embeddings as
$E_{pv}^{j}=\operatorname{IB}(I_{pv}^{j})$ and
$E_{click}^{k}=\operatorname{IB}(I_{click}^{k})$ correspondingly.
In our user interest reconstruction method,
we treat the embeddings of PV images $E_{pv}$
as queries $Q$,
and we input the embeddings of click images $E_{click}$
as values $V$ and keys $K$.
Then the user interest reconstruction layer can be
calculated by
\begin{align}
    R_{pv}^j & = \operatorname{Attention}(E_{pv}^j, E_{click}, E_{click}) \\
               & = \sum \alpha_k E_{click}^k
\end{align}
where $\alpha \sim softmax\left(E_{pv}^j E_{click}^T\right)$.
$\alpha_k$ is the attention on $k$th history click item.
The reason for its name (user intention reconstruction) is that the attention layer
forces the outputs to be a weighted sum of the embeddings of
historical click sequences.
Thus, the output space is limited to the combination of
$E_{click}$ within a simplex.

\subsection{Contrastive training method\label{sec:contrastive}}

\begin{figure*}[h]
    \includegraphics[width=\textwidth]{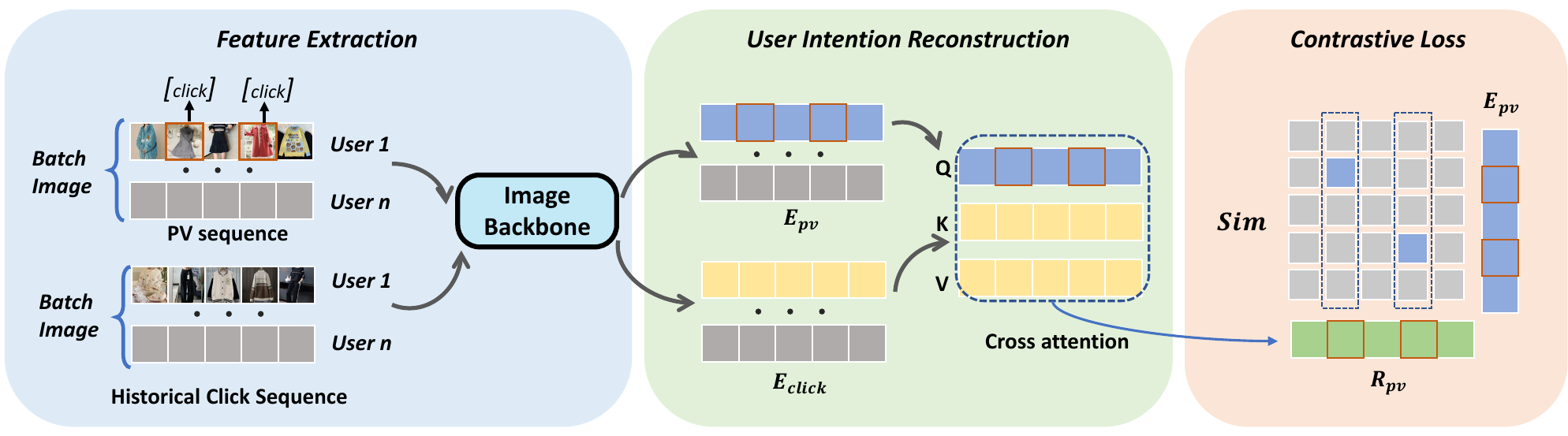}
    \caption{The contrastive user intention reconstruction method.
        The images are fed into the image backbone model to obtain the corresponding
        embeddings.
        The embeddings of PV (Page-View) sequences are blue-colored,
        and the embeddings of click sequences are yellow-colored.
        The reconstructions are in green.
        Red boxes denote positive PV items.
        \label{fig:main}}
\end{figure*}

The user intention reconstruction module cannot prevent
the trivial solution that all the embeddings collapse to
the same value.
Thus, we propose a contrastive method to train
the user-interest reconstruction module.

With the PV embeddings $E_{pv}$,
and the corresponding reconstructions $R_{pv}$.
We calculate pairwise similarity score of $E_{pv}^{j_0}$ and
$R_{pv}^{j_1}$, where $E_{pv}^{j_0}$ represents embedding of the $j_0$-th PV item and $R_{pv}^{j_1}$ represents the reconstruction of the $j_1$-th PV item:

\begin{equation}
    Sim(j_0, j_1) = \frac{{E_{pv}^{j_0}}^T R_{pv}^{j_1}}{||E_{pv}^{j_0}|| \; ||R_{pv}^{j_1}||}
\end{equation}
Then we calculate the contrastive loss by
\begin{align}
    L_{pv} & =   \mathcal{L}_{contrast}(E_{pv}, E_{click}, y)                                                                                           \\
           & =  \sum_{j_0, j_1} -\log \left(\frac{e^{Sim(j_0, j_1)/\tau}}{\sum_{j_0} e^{Sim(j_0, j_1)/\tau}}\right) \mathbb{I}[j_0=j_1 \; and \; y_{j_0}=1]
\end{align}
Here $\mathbb{I}[j_0=j_1 \; and \; y_{j_0}=1]$
is an indicator function that equals $1$ when $j_0=j_1$ and $y_{j_0}=1$,
and equals $0$ otherwise.

The contrastive loss with user interest reconstruction
is depicted in \reffig{fig:main}.
The softmax function is calculated column-wisely,
and only the positive PV images are optimized to
be reconstructed (the two columns in dashed boxes).
The behavior of the contrastive loss aligns with our assumption:
Positive PV images are pulled closer to the corresponding reconstructions ($j_0=j_1$ and $y_{j_0}=1$),
while the negative PV images are not encouraged to be reconstructed ($j_0=j_1$ and $y_{j_0}=0$).
All the PV embeddings $E_{pv}^{j_0}$ with $j_0 \ne j_1$ and $y_{j_0}=1$,
are pushed further away from $R_{pv}^{j_1}$,
which can prevent the trivial solution that all the embeddings are the same.
Some elements in the similarity matrix are not used in our loss,
namely the elements with $y_{j_1}=0$,
since the negative PV images are not supposed to be reconstructible.
We left these columns for ease of implementation. 

\noindent \textbf{Extending to batched contrastive loss}.
The above contrastive loss is calculated using PV items
within a single page (we have at most 10 items on a page),
which can only provide a limited number of negative samples.
However, a well-known property of contrastive loss
is that the performance increases as the number of
negative samples increases,
which is verified both practically\cite{simclr}
and theoretically\cite{vi_bound}.
To increase the number of negative samples,
we propose to treat all other in-batch PV items as
negative samples.
Specifically, we have (batch\_size*$l_{pv}$) PV items in this batch.
For a positive PV item, all the other (batch\_size*$l_{pv}-1$) PV items
are used as negatives, which significantly increases the number of
negatives.

\subsection{Contrastive learning on click sequences}
In the contrastive loss introduced in \refsec{sec:contrastive},
the negative and positive samples are from the same query.
Since all the PV items that are pushed to the users are already
ranked by our online recommender systems,
they may be visually and conceptually similar and
are relatively hard to distinguish by contrastive loss.
Although the batch training method introduces many
easier negatives,
the hard negatives still dominate the contrastive loss
at the beginning of training,
which makes the model hard to train.

Thus, we propose another contrastive loss on the user click history,
which does not have hard PV negatives.
Specifically, suppose the user's click history is denoted by
corresponding embeddings
$E_{click}^{0}, ..., E_{click}^{l_{click}-1}$
We treat the last item as the next item to be clicked (since the items are sorted by their click timestamp),
and the rest items are treated as click history.
The user click sequence loss is calculated as follows:
\begin{align*}
    L_{ucs} = \mathcal{L}_{contrast}(E_{click}^{l_{click}-1}, E_{click}^{0:l_{click}-1}, \mathbf{1})
\end{align*}
The $\mathcal{L}_{contrast}$ function is the same as in \refsec{sec:contrastive}.
Here the label $y=\mathbf{1}$ because all the samples in history click sequences are positive.
The user sequence loss provides an easier objective at the start of the training,
which helps the model to learn with a curriculum style.
It also introduces more signals to train the model,
which improves data efficiency.
The overall loss of \name{} is:
\begin{equation}
    L_{\name{}} = L_{pv} + L_{ucs}
\end{equation}

\subsection{Differences compared to established contrastive methods}

Although \name{} is also a contrastive method,
it is significantly different from classic contrastive methods.
First, in contrastive methods such as SimCLR\cite{simclr} and CLIP\cite{clip},
every sample has a corresponding positive counterpart.
In our method, a negative PV item does not have a corresponding positive reconstruction,
but it still serves as a negative sample in the calculation.
Secondly, there is a straightforward one-to-one matching in SimCLR and CLIP,
e.g., text and corresponding image generated by augmentation.
In recommendation systems, image augmentations are not applicable due to the distortion of appearance.
Instead, a positive PV item corresponds to a list
of history click items,
which is transformed into a one-to-one matching with our
user interest recommendation module introduced in \refsec{sec:reconstruction}.
Thirdly, another approach to convert this multi-to-one matching to one-to-one
is to apply self-attention on the multi-part, as suggested in \cite{multi_contras},
which turned out to perform worse in our scenario.
We experiment and analyze this method in \refsec{sec:abl} (w/o Reconstruction).

\section{Experiments}

To validate the efficacy of our proposed image representation learning method in enhancing recommendation performance, we conducted experiments on several publicly available product recommendation datasets. Subsequently, we proceeded with systematic experiments on actual data from Taobao and integrated our approach into the online system.\footnote{Code and Appendix are provided at \url{https://github.com/ThyrixYang/COURIER}}

\subsection{Experiments on public datasets}

Since nearly all recommendation methods incorporate ID features, the addition of image or text features typically characterizes multimodal recommendation methods. Our CTR prediction model falls under this category as well. These methods primarily rely on fixed image and text backbone models to extract image features. However, our proposed image feature pre-training method aims to enhance the representational capabilities of image features in recommendation system methods. In this section, we will experimentally investigate whether our pre-trained image features can further enhance the performance of these methods.

\noindent \textbf{Datasets}. In order to validate the effectiveness of our proposed visual feature pre-training method in learning user-intention related features, we select three categories from the Amazon Review dataset\cite{img_rec_sigir15}, namely Baby, Sports, and Clothing. Each item has corresponding image and text descriptions. We apply our method to the image features while keeping the text features unchanged.
The statistics are shown in \reftab{tab:public_datasets}.
To evaluate the performance of different visual feature extraction methods, we divided the dataset into pre-training, training, validation, and testing sets, with proportions of 50\%, 30\%, 10\%, and 10\% respectively.
We group the datasets by user IDs to construct a list of positive items, and then we uniformly sample 10 negative items for each user to construct the pre-training dataset.

\begin{table}[htbp]
  \centering
  \caption{Statistics of public datasets.\label{tab:public_datasets}}
    \begin{tabular}{lrrr}
    \toprule
          & \multicolumn{1}{l}{\# User} & \multicolumn{1}{l}{\# Items} & \multicolumn{1}{l}{\# Interactions} \\
    \midrule
    Baby  & 19445 & 7050  & 160792 \\
    Sports & 35598 & 18357 & 296337 \\
    Clothing & 39387 & 23033 & 278677 \\
    \bottomrule
    \end{tabular}%
\end{table}%

\noindent \textbf{Baselines}.
We compare with several representative ID-based and multi-modal recommendation methods.
We select two ID-based recommendation methods, namely BPR\cite{no_e2e_image_rec_b} and SLMRec\cite{slmrec}.
We select five multi-modal recommendation methods, namely DualGNN\cite{dualgnn}, LATTICE\cite{lattice},
MGCN\cite{mgcn}, MMGCN\cite{mmgcn} and BM3\cite{bm3}.
To validate the information gained from our visual feature learning method,
we concatenate the pre-trained embeddings with the original embedding provided by \citet{mmrec}.
Then, we apply the augmented embeddings to BM3 and MMGCN, which corresponds to BM3+ours and MMGCN+ours, respectively.

\noindent \textbf{Implementation}. All the baseline methods are tuned following \citet{mmrec}. Specifically, we tune the hyper-parameters (learning rate, weight decay, and method-specific hyper-parameters such as contrastive loss weight in MGCN) of each method with a validation dataset. Then, we run each method with 5 different random seeds and report their average performance.

\noindent \textbf{Results}. As observed in \reftab{tab:public}, our method, when combined with existing multimodal recommendation algorithms, can further enhance performance, achieving an average improvement of around 12\% in terms of both Recall and NDCG.

\begin{table}[htbp]
    \centering
    \caption{Averge Recall and NDCG performance comparison on public datasets.\label{tab:public}}
    \tiny
    \begin{tabularx}{0.95\textwidth}{lccccccccc}
        \toprule
        Dataset    & \multicolumn{3}{c}{Sports} & \multicolumn{3}{c}{Baby} & \multicolumn{3}{c}{Clothing}                                                                                                            \\
        \midrule
        Method     & R@20                       & R@50                     & N@50                         & R@20            & R@50            & N@50            & R@20            & R@50            & N@50           \\
        \midrule
        BPR        & 0.004                      & 0.008                    & 0.003                        & 0.007           & 0.014           & 0.005           & 0.004           & 0.007           & 0.002          \\
        SlmRec     & 0.006                      & 0.016                    & 0.006                        & 0.014           & 0.028           & 0.011           & 0.004           & 0.010           & 0.003          \\
        \midrule
        DualGNN    & 0.012                      & 0.023                    & 0.009                        & 0.018           & 0.037           & 0.014           & 0.007           & 0.014           & 0.005          \\
        LATTICE    & 0.012                      & 0.021                    & 0.008                        & 0.010           & 0.022           & 0.008           & -               & -               & -              \\
        MGCN       & 0.015                      & 0.027                    & 0.011                        & 0.017           & 0.032           & 0.013           & 0.011           & \textbf{0.019}  & \textbf{0.008} \\
        MMGCN      & 0.015                      & 0.028                    & 0.012                        & 0.024           & 0.061           & 0.024           & 0.008           & 0.017           & 0.006          \\
        BM3        & 0.018                      & 0.033                    & 0.014                        & 0.031           & 0.065           & 0.026           & 0.009           & 0.016           & 0.006          \\
        \midrule
        MMGCN+ours & 0.017                      & 0.031                    & 0.013                        & 0.030           & 0.061           & 0.024           & 0.010           & \textbf{0.019 } & 0.007          \\
        BM3+ours   & \textbf{0.019 }            & \textbf{0.034 }          & \textbf{0.015 }              & \textbf{0.035 } & \textbf{0.068 } & \textbf{0.027 } & \textbf{0.012 } & \textbf{0.019 } & 0.007          \\
        \bottomrule
    \end{tabularx}
\end{table}%

\subsection{Experiments on Taobao offline dataset}

To evaluate the information increment of pre-trained image embeddings on
the CTR prediction task,
we use an architecture that aligns with our online recommender system.
In essence, each item is associated with a unique item ID and features. The item ID is transformed into an embedding and concatenated with other features, including the image embedding we have learned, to form the item features. Users consist of user features and their behavioral history, where the behavioral history comprises numerous items. Therefore, the user behavioral history is aggregated using self-attention, which is then concatenated with other user features. Subsequently, the user and item features are concatenated and serve as inputs to an MLP. The MLP's final output is the user's click probability, representing the predicted CTR.
We provide a detailed description of this CTR model in the Appendix.

\noindent \textbf{Downstream usage of image representations}.
Practically, we find that how we feed the image features
into the downstream CTR model is critical for the final performance.
We experimented with three different methods: 1. directly using embedding vectors.
2. using similarity scores to the target item.
3. using the cluster IDs of the embeddings.
Cluster-ID is the best-performing method among the three methods,
bringing about $0.1\%-0.2\%$ improvements on AUC compared to
using embedding vectors directly.
We attribute the success of Cluster-ID to its better alignment with
our pre-training method. We provide a more detailed analysis in the Appendix.

\subsubsection{Taobao pre-training and CTR dataset\label{sec:exp_datasets}}

\begin{table}[htbp]
    \centering
    \caption{Pre-training dataset collected from the women's clothing category.}
    \begin{tabular}{llllrr}
        \toprule
        \# User      & \# Item      & \# Samples    & CTR  & \multicolumn{1}{l}{\# Hist} & \multicolumn{1}{l}{\# PV Items} \\
        \midrule
        71.7 million & 35.5 million & 311.6 million & 0.13 & 5                           & 5                               \\
        \bottomrule
    \end{tabular}%
    \label{tab:train_dataset}%
\end{table}%

\noindent \textbf{Pre-training dataset.} The pre-training dataset is collected during
2022.11.18-2022.11.25 on our online search service.
To reduce the computational burden,
we down-sample to 20\% negative samples.
So the click-through rate (CTR) is increased to around 13\%.
To reduce the computational burden,
we sort the PV items with their labels to list positive samples in the front,
then we select the first 5 PV items to constitute the training targets ($l_{pv}=5$).
We retain the latest 5 user-click-history items ($l_{click}=5$).
Thus, there are at most 10 items in one sample.
There are three reasons for such data reduction:
First, our dataset is still large enough after reduction.
Second, the number of positive items in PV sequences is less than 5 most of the time,
so trimming PV sequences to 5 will not lose many positive samples,
which is generally much more important than negatives.
Third, we experimented with $l_{click}=10$ and did not observe
significant improvements, while the training time is significantly longer.
Thus, we keep $l_{click}=5$ and $l_{pv}=5$ fixed in all the experiments.
We remove the samples without clicking history or positive items within the page.
Intuitively, women's clothing is one of the most
difficult recommendation tasks (the testing AUC is significantly lower than average)
which also largely depends on the visual appearance of the items.
Thus, we select the women's clothing category to form the dataset of
the training dataset of the pre-training task.
The statistics of the pre-training dataset are
summarized in \reftab{tab:train_dataset}.

In the pre-training dataset,
we only retain the item images and the click labels.
All other information, such as item title,
item price, user properties, and even the query by the users
are dropped.
We report the experimental results of training with text information in
\refsec{sec:train_with_clip}, which indicates that additional information is unnecessary.

\begin{table}[h]
    \centering
    \caption{Daily average statistics of the downstream dataset.}
    \begin{tabular}{llllrr}
        \toprule
                & \# User       & \# Item       & \# Samples     & \multicolumn{1}{l}{CTR} & \multicolumn{1}{l}{\# Hist} \\
        \midrule
        All     & 0.118 billion & 0.117 billion & 4.64 billion   & 0.139                   & 98                          \\
        Women's & 26.39 million & 12.29 million & 874.39 million & 0.145                   & 111.3                       \\
        \bottomrule
    \end{tabular}%
    \label{tab:test_dataset}%
\end{table}%

\noindent \textbf{CTR dataset.}
The average daily statistics of the downstream CTR datasets in all categories and women's clothing are summarized in \reftab{tab:test_dataset}.
To avoid any information leakage,
we use data collected from 2022.11.27 to 2022.12.04 on our online shopping platform to train the downstream
CTR model,
and we use data collected on 2022.12.05 to evaluate the performance.
In the evaluation stage,
the construction of the dataset aligns with our online system.
We use all the available information to train the downstream
CTR prediction model.
The negative samples are also down-sampled to 20\%.
Different from the pre-training dataset,
we do not group the page-view data in the evaluation dataset,
so each sample corresponds to an item.

\subsubsection{Evaluation metrics\label{sec:metrics}}

\begin{itemize}
    \item Area Under the ROC Curve (\textbf{AUC}):
          AUC is the most commonly used evaluation metric
          in evaluating ranking methods, which denotes
          the probability that a random positive sample is ranked
          before a random negative sample.

    \item Grouped AUC (\textbf{GAUC}):
          The AUC is a global metric that ranks all the predicted
          probabilities.
          However, in the online item-searching task,
          only the relevant items are considered by the
          ranking stage,
          so the ranking performance among the recalled items (relevant to the user's query) is more meaningful than
          global AUC.
          Thus, we propose a Grouped AUC metric, which is the average AUC
          within searching sessions.
\end{itemize}

\subsubsection{Compared methods}

\begin{itemize}
    \item \textbf{Baseline}:
          Our baseline is a CTR model that serves our online system.
          It's noteworthy that we adopt a warmup strategy that uses our online model trained with
          more than one year's data to initialize
          all the weights (user ID embeddings, item ID embeddings, etc.),
          which is a fairly strong baseline.
    \item \textbf{Supervised}:
          To pre-train image embeddings with user behavior
          information,
          a straightforward method is to train a CTR model with
          click labels and image backbone network end-to-end.
          We use the trained image network to extract embeddings
          as other compared methods.

    \item \textbf{SimCLR}\cite{simclr}:
          SimCLR is a self-supervised image pre-training method based
          on augmentations and contrastive learning.

    \item \textbf{SimSiam}\cite{simsiam}:
          SimSiam is also an augmentation-based method.
          Different from SimCLR,
          SimSiam suggests that contrastive loss is unnecessary and
          proposes directly minimizing the distance between matched embeddings.

    \item \textbf{CLIP}\cite{clip}:
          CLIP is a multi-modal pre-training method that
          optimizes a contrastive loss between image embeddings
          and item embeddings.
          We treat the item cover image and its title as a matched sample.
          We use a pre-trained BERT\cite{chinese_bert} as the feature network of item titles,
          which is also trained end-to-end.
    \item \textbf{MaskCLIP}\cite{mask-clip}:
        MaskCLIP is an improved version of CLIP with masked self-distillation in images and text.
\end{itemize}

\subsubsection{Performance in downstream CTR task\label{sec:perform_ctr}}

\begin{table*}[h]
    \centering
    \caption{
        The improvements of AUC ($\Delta \text{AUC}$) in the women's clothing category.
        And performances of $\Delta \text{AUC}, \Delta \text{GAUC}$ in all categories.
        We report the relative improvements compared to the Baseline method,
        and the raw values of the metrics are in parentheses.
    }
    \resizebox{\linewidth}{!}{%
        \begin{tabular}{lrrrr}
            \toprule
            Methods               & \multicolumn{1}{l}{$\Delta \text{AUC}$ (Women's Clothing)} & \multicolumn{1}{l}{$\Delta \text{AUC}$}   & \multicolumn{1}{l}{$\Delta \text{GAUC}$}  \\
            \midrule
            Baseline              & 0.00\% (0.7785)                            & 0.00\% (0.8033)           & 0.00\% (0.7355)           \\
            Supervised            & +0.06\% (0.7790)                           & -0.14\% (0.8018)          & -0.06\% (0.7349)          \\
            CLIP\cite{clip}       & +0.26\% (0.7810)                           & +0.04\% (0.8036)          & -0.09\% (0.7346)          \\
            SimCLR\cite{simclr}   & +0.28\% (0.7812)                           & +0.05\% (0.8037)          & -0.08\% (0.7347)          \\
            SimSiam\cite{simsiam} & +0.10\% (0.7794)                           & -0.10\% (0.8022)          & -0.29\% (0.7327)          \\
            MaskCLIP\cite{mask-clip} & +0.31\% (0.7815)                           & +0.03\% (0.8035)          & -0.03\% (0.7352)          \\
            \name(ours)           & \textbf{+0.46\% (0.7830)}                  & \textbf{+0.16\% (0.8048)} & \textbf{+0.19\% (0.7374)} \\
            \bottomrule
        \end{tabular}%
    }
    \label{tab:main}%
\end{table*}%

The performances of compared methods are summarized in \reftab{tab:main}.
We have the following conclusions:
First, since all the methods are pre-trained in the women's clothing category,
they all have performance improvement on the AUC of the downstream women's clothing category.
SimCLR, SimSiam, CLIP, MaskCLIP and \name{} outperform the Baseline and Supervised pre-training.
Among them, our \name{} performs best, outperforms baseline by $0.46\%$ AUC and
outperforms second best method MaskCLIP by $0.15\%$ AUC, which verified our analysis that traditional CV pre-training methods provide little information gain to the CTR model.
Secondly, we also check the performance in all categories.
Our \name{} also performs best with $0.16\%$ improvement in AUC.
However, the other methods' performances are significantly different
from the women's clothing category.
The Vanilla and SimSiam method turned out to have a negative impact in all categories.
And the improvements of CLIP, SimCLR and MaskCLIP become marginal.
The reason is that the pre-training methods failed to extract general user interest information
and overfit the women's clothing category.
We analyze the performance in categories other than women's clothing in \refsec{sec:cate_auc}.
Thirdly, the performance on GAUC indicates that the performance gain of CLIP, SimCLR and MaskCLIP
vanishes when we consider in-page ranking,
which is indeed more important than global AUC, as discussed in \refsec{sec:metrics}.
The GAUC performance further validates that \name{} can learn fine-grained
user interest features that can distinguish between in-page items.

\subsection{Further experimental analysis}

We have validated the efficacy of the COURIER approach in enhancing the overall performance of recommendation algorithms on both public datasets and the Taobao dataset. However, we also aim to address the following inquiries: 

\begin{enumerate}
    \item Existing research has highlighted the significant impact of temperature coefficients in pre-training. Does the temperature coefficient similarly affect our pre-training task?
    \item What is the respective contribution of each module in the COURIER approach towards performance improvement?
    \item Currently, we only utilize image and user click information in the training of embeddings. Would the inclusion of other modalities, such as text, further enhance the performance?
    \item As a pre-training method, can the embeddings acquired in the women's clothing category also yield improvements in other categories?
    \item Can our method solely utilize user click information to learn features relevant to user intent?
\end{enumerate}

We will address the aforementioned inquiries through experimental investigation.

\subsubsection{Influence of temperature}

\begin{figure}
    \centering
    \includegraphics[width=0.7\linewidth]{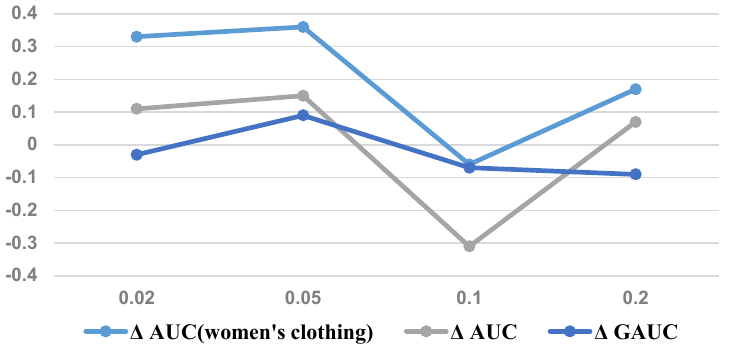}
    \caption{
          The impact of different values of temperature $\tau$ on the performance of downstream CTR tasks. 
          The horizontal axis represents the values of $\tau$, 
          while the vertical axis denotes the change (\%) in the metrics.
        \label{fig:influence_tau}
    }
\end{figure}

We experiment with different $\tau$. The results are in \reffig{fig:influence_tau}.
Note that the experiments are run with the Simscore method,
which is worse than Cluster-ID but is much faster.
We find that $\tau=0.05$ performs best for \name{} and keep it fixed.

\subsubsection{Ablation study\label{sec:abl}}

\begin{table}[htbp]
    \centering
    \caption{Ablation studies of \name.}
    \begin{tabular}{lrrr}
        \toprule
                           & \multicolumn{1}{l}{$\Delta \text{AUC}$ (women's clothing)} & \multicolumn{1}{l}{$\Delta \text{AUC}$} & \multicolumn{1}{l}{$\Delta \text{GAUC}$} \\
        \midrule
        w/o UCS            & 0.06\%                                     & -0.13\%                 & 0.11\%                   \\
        w/o Contrast       & 0.23\%                                     & 0.03\%                  & -0.11\%                  \\
        w/o Reconstruction & 0.25\%                                     & 0.02\%                  & -0.11\%                  \\
        w/o Neg PV         & 0.30\%                                     & 0.07\%                  & -0.06\%                  \\
        \name              & \textbf{0.46\%}                            & \textbf{0.16\%}         & \textbf{0.19\%}          \\
        \bottomrule
    \end{tabular}%
    \label{tab:ablation}%
\end{table}%

We conduct the following ablation experiments to verify
the effect of each component of \name{}.
\begin{itemize}
    \item \textbf{w/o UCS}: Remove the user click sequence loss.
    \item \textbf{w/o Contrast}: Remove the contrastive loss, only minimize the
          reconstruction loss, similar to SimSiam\cite{simsiam}.
    \item \textbf{w/o Reconstruction}: Use self-attention instead of cross-attention
          in the user interest recommendation module.
    \item \textbf{w/o Neg PV}: Remove negative PV samples, only use positive samples.
\end{itemize}
The results in \reftab{tab:ablation} indicate that all the
proposed components are necessary for the best performance.

\subsubsection{Influence of batch size}

Due to both theoretical\cite{vi_bound} and practical evidence\cite{simclr} indicating that increasing batch size in contrastive learning can enhance model generalization, we conducted experiments with different batch sizes.
The results in \reftab{tab:batch_size} also demonstrate that in our method, the larger the batch size, the better the performance.

On the flip side, augmenting the batch size necessitates a substantial increase in computational power. In our framework, configuring a batch size of 3072 requires 48 Nvidia V100 GPUs, while a batch size of 4096 demands 64 GPUs. Considering the diminishing returns associated with further escalating the batch size, coupled with the consideration of training costs, we ultimately opted for a batch size of 3072 for deployment.

\begin{table}[htbp]
  \centering
  \caption{
    Influence of different batch on performance\label{tab:batch_size}
  }
    \begin{tabular}{rrrr}
    \toprule
    \multicolumn{1}{l}{Batch Size} & \multicolumn{1}{l}{Femal AUC} & \multicolumn{1}{l}{AUC} & \multicolumn{1}{l}{GAUC} \\
    \midrule
    64    & 0.15\% & -0.06\% & -0.08\% \\
    256   & 0.23\% & 0.04\% & 0.02\% \\
    512   & 0.36\% & 0.09\% & 0.10\% \\
    2048  & 0.43\% & 0.15\% & 0.17\% \\
    3072  & 0.46\% & 0.16\% & 0.19\% \\
    4096  & 0.47\% & 0.16\% & 0.21\% \\
    \bottomrule
    \end{tabular}%
\end{table}%

\subsubsection{Train with text information\label{sec:train_with_clip}}

\begin{table}[h]
    \centering
    \caption{Train with text information.}
    \begin{tabular}{lrrr}
        \toprule
               & \multicolumn{1}{l}{$\Delta \text{AUC}$ (women's clothing)} & \multicolumn{1}{l}{$\Delta \text{AUC}$} & \multicolumn{1}{l}{$\Delta \text{GAUC}$} \\
        \midrule
        w CLIP & 0.26\%                                     & 0.04\%                  & -0.09\%                  \\
        \name  & \textbf{0.46\%}                            & \textbf{0.16\%}         & \textbf{0.19\%}          \\
        \bottomrule
    \end{tabular}%
    \label{tab:add_clip}%
\end{table}%

Text information is important for searching and recommendation
since it's directly related to the query by the users and the
properties of items.
Thus, raw text information is already widely used in real-world systems.
The co-train of texts and images also shows significant performance
gains in computer vision tasks such as classification and segmentation.
So we are interested in verifying the influence to \name{} by co-training with text
information.
Specifically, we add a CLIP\cite{clip} besides \name{},
with the loss function becomes $L=L_{\name{}} + L_{CLIP}$.
The CLIP loss is calculated with item cover images and
item titles.
However, such multi-task training leads to worse downstream CTR performance
as shown in \reftab{tab:add_clip},
which indicates that co-training with the text information may
not help generalization when the text information is available in
the downstream task.

\subsubsection{Generalization in unseen categories\label{sec:cate_auc}}

\begin{figure}[h]
    \centering
    \includegraphics[width=0.9\linewidth]{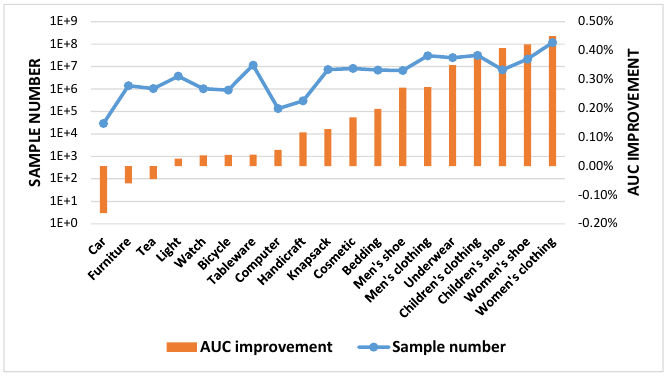}
    \caption{
        The AUC improvements of \name{} compared to the Baseline on
        different categories.
        The x-axis is sorted by the improvements.
        \label{fig:cate_compare}
    }
\end{figure}

In \reffig{fig:cate_compare}, we plot the AUC improvements of \name{} in different categories.
We have the following conclusions:
First, the performance improvement in the women's clothing category is
the most, which is intuitive since the embeddings are trained
with women's clothing data.
Secondly, there are also significant improvements in
women's shoe, children's shoe, children's clothing, underwear, etc.
These categories are not used in the pre-training task,
which indicates the \name{} method can learn general
visual characteristics that reflect user interests.
Thirdly, the performances in the bedding, cosmetics, knapsack,
and handicrafts are also improved by more than 0.1\%.
These categories are significantly different from
women's clothing in visual appearance,
and \name{} also learned some features that are transferable to
these categories.
Fourthly, \name{} does not have a significant impact on
some categories, and has a negative impact on the car category.
These categories are less influenced by visual looking and
can be ignored when using our method to avoid performance drop.
Fifthly, the performance is also related to the amount of data.
Generally, categories that have more data tend to perform better.

\subsubsection{Visualization of trained embeddings}

\begin{figure*}
    \includegraphics[width=\textwidth]{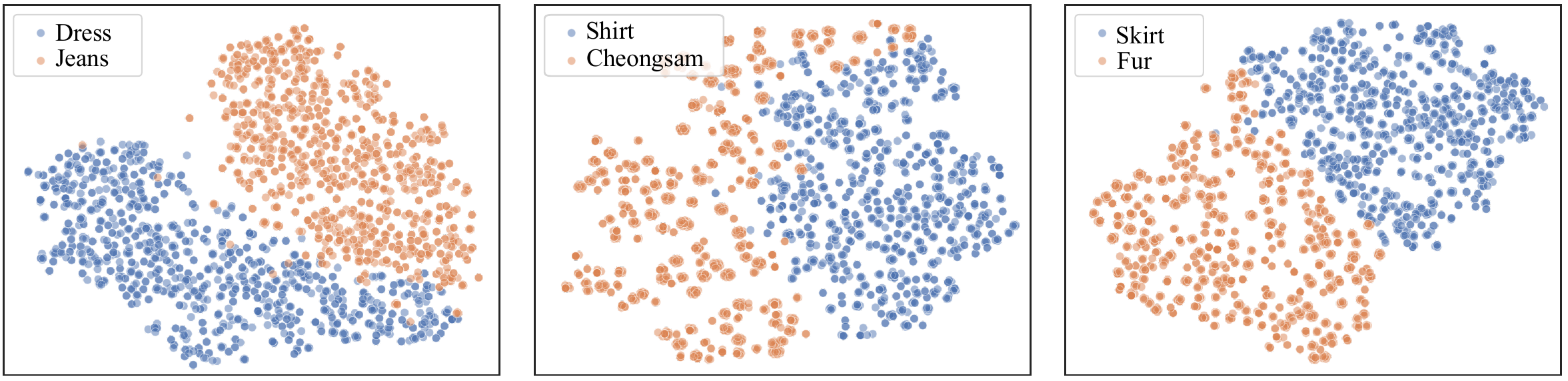}
    \caption{T-SNE visualization of embeddings in different categories.\label{fig:viz_cate}}
\end{figure*}

\begin{figure*}
    \includegraphics[width=\textwidth]{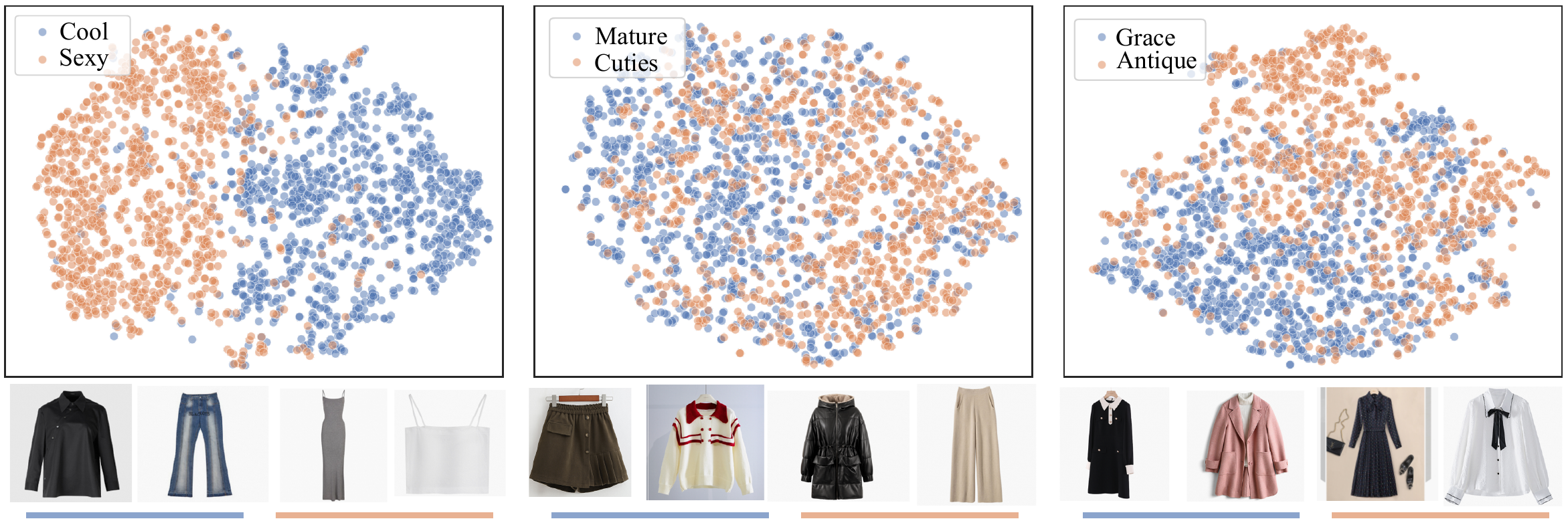}
    \caption{T-SNE visualization of embeddings with different style tags.
        We also plot some item images with different tags below the corresponding figures.
        \label{fig:viz_style}}
\end{figure*}

Did \name{} really learn features related to user interests?
We verified the quantitative improvements on CTR in \refsec{sec:perform_ctr}.
Here, we also provide some qualitative analysis.
During the training of \name{},
we \textit{did not} use any additional information other than images and user clicks.
Thus, if the embeddings contain some semantic information,
such information must be extracted from user behaviors.
So we plot some randomly selected embeddings with specific categories and style tags in
\reffig{fig:viz_cate} and \reffig{fig:viz_style}.
First, embeddings from different categories are clearly
separated, which indicates that \name{} can learn categorical
semantics from user behaviors.
Secondly, some of the style tags can be separated, such as Cool vs. Sexy.
The well-separated tags are also intuitively easy to distinguish.
Thirdly, some of the tags can not be separated clearly, such as Mature vs. Cuties, and Grace vs. Antique,
which is also intuitive since these tags have relatively vague meanings and
may overlap.
Despite this, \name{} still learned some gradients between the two concepts.
To conclude, the proposed \name{} method can learn
meaningful user-interest-related features by only using images and click labels.

\section{Online experiments and deployment}

During the deployment of our image feature learning method, we devoted substantial efforts to optimizing the model's performance. Below, we summarize some of the impactful optimization measures that were undertaken:

\noindent \textbf{Image backbone and performance optimization}.
We use the Swin-tiny transformer\cite{swin} trained on Imagenet as the image backbone model,
which is faster than the Swin model and has comparable performance.
We train the backbone network end-to-end in the pre-training phase.
However, since we have about $3\times 10^9$ images in our pre-training dataset
(230 times bigger than the famous Imagenet dataset, which has about $1.3\times 10^6$ images),
even the Swin-tiny model is slow.
After applying gradient checkpointing\cite{gradient_checkpoint},
mixed-precision computation\cite{mixed_precision},
and optimization on distributed IO of images,
we managed to reduce the training time for one epoch on the pre-training
dataset from ~150 hours to ~50 hours (consuming daily data in $<$ 10 hours)
with 48 Nvidia V100 GPUs (32GB).

\noindent \textbf{Efficient clustering}.
In our scenario, there are about $N=6*10^7$ image embeddings to be clustered,
and we set the cluster number to $C=10^5$.
The computational complexity of vanilla k-means implementation is $O(N*C*d)$ per iteration,
which is unaffordable.
Practically, we implement a high-speed learning-based clustering problem proposed in \cite{fast_cluster}.
The computing time is reduced significantly from more than 7 days to about 6 hours.
We will assign a new image to its closest cluster center ID. The amount of new items is relatively small and has little impact on the performance. We will re-train and replace all the cluster-IDs regularly (e.g., half a year).

\begin{table}[htbp]
    \centering
    \caption{The A/B testing improvements of \name{}.}
    \begin{tabular}{llll}
        \toprule
                             & $\Delta$ \# Order & $\Delta$ CTR     & $\Delta$ GMV     \\
        \midrule
        All categories    & +0.1\%   & +0.18\% & +0.66\% \\
        Women's clothing & +0.31\%  & +0.34\% & +0.88\% \\
        \bottomrule
    \end{tabular}%
    \label{tab:abtest}%
\end{table}%

\noindent \textbf{Online serving performance}. To evaluate the performance improvement brought by \name{} on our online system,
we conduct online A/B testing in our online shopping platform for 30-days.

We report improvements on number of orders ($\Delta$\#Order), click-through rate ($\Delta$ CTR), and cross merchandise volume ($\Delta$ GMV).
Compared with the strongest deployed online baseline,
COURIER significantly (p-value $<$ 0.01) improves the CTR and GMV by +0.34\%
and +0.88\% in women's clothing, respectively (the noise level is less than 0.1\% according to the online A/A test).
Such improvements are considered significant with the
large volume of our online shopping platform.
The model has also been successfully deployed into production,
serving the main traffic.

\section{Conclusion}

Visually, the image information of a product has a significant impact on whether a user clicks. However, we have observed that the features extracted by existing image pre-training methods have limited utility for improving downstream CTR models. We attribute this phenomenon to the fact that the labels typically used in existing methods, such as those for classification or image-text retrieval, have already been applied as features in CTR models. Existing image pre-training methods are insufficient for effectively extracting features relevant to user interests.
To address this issue, we propose a method for user interest reconstruction to extract image features relevant to user click behavior. Specifically, to mitigate the problem of images in a user's click history that may not be relevant to the current image, we employ an attention-based approach to identify images in the click sequence that are similar to the current image. We then attempt to reconstruct the user's next-click image through a weighted sum of the embeddings of these related images. Furthermore, to prevent trivial solutions, we optimize using a contrastive learning loss, reducing the reconstruction error for clicked images while increasing it for non-clicked images. Consequently, our method learns visual embeddings that influence whether a user clicks without relying on any downstream features directly. Instead, it enhances the information available for downstream utilization from the perspective of interest reconstruction.
Experiments conducted on various datasets and large-scale online evaluations confirm that the embeddings learned by our method significantly enhance the performance of downstream recommendation models.

\bibliography{sn-bibliography}%

\begin{appendices}
\section{CTR model\label{sec:ctr_model}}

The CTR model consists of a user embedding network,
an item embedding network, and a query embedding network.

\begin{figure*}[h]
    \includegraphics[width=\textwidth]{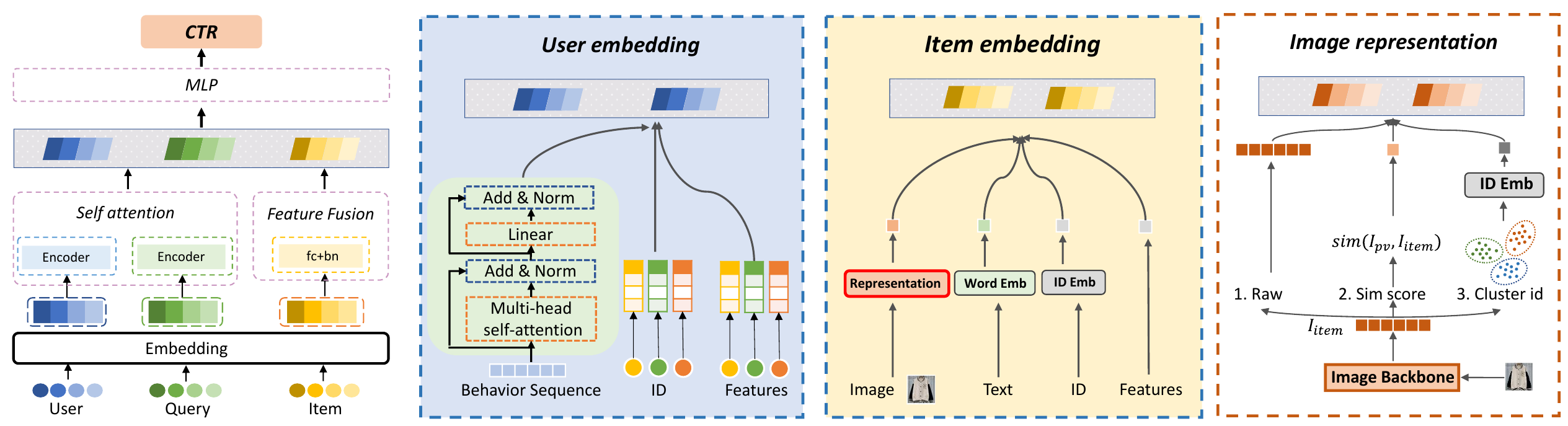}
    \caption{The downstream CTR model and image representation.\label{fig:ctr_model}}
\end{figure*}

\noindent \textbf{Item embedding network}.
The item embedding network takes the
cover image of the item, the item ID, the item title,
and various statistical features as input,
as depicted in \reffig{fig:ctr_model}.
The image features are fed by one of the three methods (Vector, Similarity score, Cluster ID).
ID features are transformed into corresponding embeddings.
The item titles are tokenized, then the tokens are converted to embeddings.
All the features are then concatenated to form the item embeddings.

\noindent \textbf{User embedding network}.
The user embeddings consist of the embeddings of history item sequences,
the user IDs, and other statistical features.
The user IDs and statistical features are treated similarly to item features.
The most important feature for personalized recommendation is the
history item sequence.
In our CTR model, we use three different item sequences:
1. Long-term click history consists of the latest up to 1000 clicks on the concerned
category (the one the user is searching for) within 6 months.
2. Long-term payment history consists of up to 1000 paid items within 2 years.
3. The recent up to 200 item sequences in the current shopping cart.
All the items are embedded by the item embedding network.
The embedded item sequences are fed to multi-head attention and layer-norm layers.
Then, the item embeddings are mean-pooled and projected to a proper dimension,
which is concatenated with other user features to form the final user embeddings.

\noindent \textbf{Query embedding and CTR prediction network}.
The user queries are treated in the same way as item titles.
The user embeddings, item embeddings, and query embeddings are
flattened and concatenated into a single vector.
Then, we use an MLP model with 5 layers to produce the logits for
CTR prediction.
The CTR model is trained with the cross entropy loss on the downstream
user click data.

\section{Downstream usage of image embeddings\label{sec:emb_usage}}

How the features are fed to the model is a critical factor
that affects the performance of machine learning algorithms.
For example, normalizing the inputs before input to neural networks
is a well-known preprocessing that matters a lot.
In our practice of pre-training and utilizing pre-trained embeddings in downstream
tasks, we also find that the way we insert the pre-trained embeddings is
critical to the downstream performance.
We explore three different approaches: Vector, Similarity score, and Cluster ID.

\textbf{Vector}.
The most straightforward and common practices of utilizing pre-trained
embeddings are to use the embeddings as input to the downstream tasks directly.
In such cases, the features of the item images are represented as embedding vectors,
which is also the first approach we have tried.
However, our experiments show that no matter how we train the image embeddings,
the improvements in the CTR task are only marginal.
We think the reason is that the embedding vectors are relatively
hard to be used by the downstream model,
so the downstream model ignores the embedding vectors.
The existing recommender systems already use item IDs as features,
and the embeddings of the IDs can be trained directly.
So the IDs are much easier to identify an item than
image embeddings,
which require multiple layers to learn to extract related information.
Thus, the model will achieve high performance by updating ID embeddings
before the image-related weights can learn useful representations for CTR.
Then the gradient vanishes because the loss is nearly converged, 
and the image-related weights won't update significantly anymore,
which is also observed in our experiments.

\textbf{Similarity Score}.
With the hypothesis about the embedding vectors,
We experiment with a much-simplified representation of the embedding vectors.
Specifically,
assuming we want to estimate the CTR of an item with $Img_{pv}$ and
a user of click history $Img_{click}^k$.
The vector approach uses $Emb_{click}^k$ and $Emb_{pv}$ as inputs directly.
Instead, we calculate a cosine similarity score for each click history items,
\begin{equation}
  sim(Img_{pv}, Img_{click}^k) = \frac{Img_{pv}^T Img_{click}^k}{||Img_{pv}|| \; ||Img_{click}^k||}
\end{equation}
and the scores (each image corresponds to a real-valued score) are used as image features.
The results are in \reftab{tab:emb_usage}.
Experimental results indicate that the simple similarity score
features perform significantly better than inserting embedding vectors
directly, although the embedding vectors contain much more information.
The performance of the similarity score method verified our hypothesis that 
embedding vectors may be too complex to be used in the CTR task.

\begin{table}[htbp]
  \caption{
    Performance of inserting image information with Vector, SimScore, and Cluster ID.
    Since we performed this comparison in the early stage of our development, 
    the exact configurations of each version are hard to describe in detail.
    And the different versions may not be comparable to each other 
    (different training data sizes, learning rates, training methods, etc.).
    We only list the version number for clarity.
    Results within each row are comparable since they are
    generated from the same version of embeddings.
    The Baseline does not use images.
    "-" denotes that we did not evaluate Cluster-ID of these versions.
    \label{tab:emb_usage}
  }
  \begin{tabular}{lrrr}
    \toprule
             & \multicolumn{1}{l}{Vector} & \multicolumn{1}{l}{SimScore} & \multicolumn{1}{l}{Cluster ID} \\
    \midrule
    Baseline & 0.00\%                     & 0.00\%                       & 0.00\%                         \\
    V1       & 0.07\%                     & 0.14\%                       & 0.23\%                         \\
    V2       & 0.08\%                     & 0.16\%                       & 0.23\%                         \\
    V3       & 0.09\%                     & 0.18\%                       & 0.28\%                         \\
    V4       & 0.06\%                     & 0.16\%                       & 0.22\%                         \\
    V5       & 0.00\%                     & 0.11\%                       &   -                             \\
    V6       & -0.02\%                    & 0.07\%                       &   -                            \\
    V7       & -0.04\%                    & -0.01\%                      &   -                             \\
    V8       & 0.04\%                     & 0.13\%                       &   -                             \\
    V9       & 0.05\%                     & 0.09\%                       &   -                             \\
    \bottomrule
  \end{tabular}%
\end{table}%

\textbf{Cluster IDs}.
Our experiments on embedding vectors and similarity scores
indicate that embedding vectors are hard to use and
simply similarity scores contain information that
can improve downstream performance.
Can we insert more information than similarity scores
but not too much?
ID features are the most important features used in
recommender systems,
and it has been observed that ID features are easier to
train than non-ID features,
and perform better\cite{one_epoch}.
Thus, we propose to transform embedding vectors into ID features.
A straightforward and efficient method is to hash
the embedding vectors and use the hash values as IDs.
However, hashing cannot retain semantic relationships
between trained embeddings,
such as distances learned in contrastive learning.
Thus, we propose to use the cluster IDs to represent
the embedding vectors instead.
Specifically,
we run a clustering algorithm\cite{fast_cluster} on all the embedding vectors,
then we use the ID of the nearest cluster center as
the ID feature for each embedding vector.
There are some benefits of using cluster IDs:
First, the clustering algorithm is an efficient approximation method that
are scalable to large data.
Second, the cluster centers learned by the clustering algorithm
have clear interpretations and retain the global and local
distance structures.
Third, since we are using Euclidean distance in the clustering algorithm,
the learned cluster IDs can retain most of the distance information
learned in the contrastive pre-training stage. 

From \reftab{tab:emb_usage}
we can conclude that the Cluster ID method consistently outperforms the Vector and SimScore method by a significant gap.
The overall trend of SimScore and Cluster ID are similar,
but evaluating with SimScore is much faster.
Thus, practically we use the SimScore method to roughly compare different hyper-parameters,
and the most promising hyper-parameters are further evaluated with the cluster ID method.

\textbf{Why Cluster-ID is suitable for contrastive pre-training methods}.
The cosine similarity is related with the Euclidean distance as follows:
\begin{equation}
  Sim(\vx, \vy) = \frac{\vx^T \vy}{||\vx|| \; ||\vy||} = \frac{||\vx||^2 + ||\vy||^2 - ||\vx - \vy||^2}{2||\vx|| \; ||\vy||}
\end{equation}
If we constrain the embedding vectors to be $\ell 2$ normalized,
i.e., $||\vx||=||\vy||=1$, which is the same as in contrastive learning.
Then we have
\begin{equation}
  Sim(\vx, \vy) = \frac{2-||\vx-\vy||^2}{2}
\end{equation}
Thus, maximizing the cosine similarity is equivalent to
minimizing the Euclidean distance between $\ell 2$ normalized
vectors.
Since we are optimizing the cosine similarity of the
embedding vectors,
we are indeed optimizing their Euclidean distances.
And such distance information is retained by 
the clustering algorithm using Euclidean distances.
By adjusting the number of clusters,
we can also change the information to be retained.
To conclude, the Cluster-ID method aligns with both the pre-training and 
downstream stages, resulting in better performance.

\section{Why not use random masked prediction and self-attention?\label{sec:why_not_mask_pred}}
Random masked prediction is a popular pre-training style,
which may also be applied to our framework.
However, considering the data-generating process, random masking is indeed
unnecessary and may cause some redundant computation and information leakage.
Because of the time series nature of user click history,
we will train on all the possible reconstruction targets in click history.
That is, we will train with (B; A) at first,
where (A) is the history and B is the target.
After the user has clicked on item C, we will train on (C; B, A), and so on.
Thus, every item in the click history will be treated as a target exactly once.

Suppose we adopted the random masking method, there is a chance that we train on
(A; C, B), then we train on (C; A, B).
Since the model has already observed the co-occurrence of A, B, and C,
the next (C; A, B) prediction will be much easier.
The random masking method also violates the potential causation that
observing A and B causes the user's interest in C.

Masked predictions typically use self-attention instead of
cross attention with knowledge about the target as our proposed
user intention reconstruction method.
We also try with the self-attention method in the experiment
section, which performs worse than our method.
The reason is that learning to predict the next item with
only a few click history is tough,
but training to figure out how the next item can be reconstructed
with the clicking history is much easier and can provide more
meaningful signals during training.

\section{Does projection head help?}

\begin{table}[htbp]
  \centering
  \caption{Adding projection to \name.}
  \begin{tabular}{lrrr}
    \toprule
                 & \multicolumn{1}{l}{AUC (women's clothing)} & \multicolumn{1}{l}{AUC} & \multicolumn{1}{l}{GAUC} \\
    \midrule
    w projection & 0.29\%                                     & 0.06\%                  & -0.04\%                  \\
    \name        & \textbf{0.46\%}                            & \textbf{0.16\%}         & \textbf{0.19\%}          \\
    \bottomrule
  \end{tabular}%
  \label{tab:extension}%
\end{table}%
Most of the contrastive pre-training methods are equipped with
projection heads.
A projection head refers to one or multiple layers
(typically an MLP) that is appended after the last embedding layer.
During training, the contrastive loss (e.g., InfoNCE) is
calculated with the output of the projection head instead of
using the embeddings directly.
After training, the projection heads are dropped,
and the embeddings are used in downstream tasks, e.g, classification.
It is widely reported that training with projection heads
can improve downstream performance\cite{simclr,simclr_v2,simsiam}.
However, the reason for the success of the projection head is still unclear.
We experiment with projection heads as an extension of \name{}, and
the results are in \reftab{tab:extension}.
We find that projection heads have a negative impact on \name{}.
There are two possible reasons:
1. In those contrastive methods, the downstream usage and the
pre-training stage are inconsistent.
In pre-training, the contrastive loss pushes embeddings and their
augmented embeddings to be closer,
which wipes away detailed information other than distance,
while the distance itself cannot be used in classification.
By adding projection heads, the projection head can learn the
distance information without wiping away much detailed information.
2. In our \name{} method,
we use the distance information learned in the pre-training stage directly
by calculating the similarity score or cluster ID.
Thus, the contrastive loss is consistent with the downstream usage.
If we train with a projection head, the embeddings are not well
suited for calculating the similarity scores since the embeddings
are not trained to learn cosine similarity directly.

\section{Implementation and Reproduction}

\textbf{Image Backbone}. 
We adopt the swin-tiny implementation in Torchvision.
The output layer of the backbone model is replaced with
a randomly initialized linear layer,
and we use weights pre-trained on the ImageNet dataset
to initialize other layers.
We apply gradient checkpoint\cite{gradient_checkpoint} on all the attention layers in swin-tiny
to reduce memory usage and enlarge batch size.
We apply half-precision (float 16) computation\cite{mixed_precision} in all our models to accelerate 
computing and reduce memory. 
We train all the methods on a cluster with 48 Nvidia V100 GPUs (32GB),
which enables single GPU batch size = 64, overall batch size = 64*48 = 3072.
Note that each line of data contains 10 images, 
so the number of images within a batch is 30720 (some are padded zeros).
The images are resized, cropped, and normalized before feeding to the backbone model.
The input image size is 224.

We use the following hyperparameters for all the methods. 
Learning rate=1e-4, embedding size=256, weight decay=1e-6.
We use the Adam optimizer.
We have tuned these hyperparameters roughly but did not 
find significantly better choices on specific methods.

\textbf{Hyperparameters of \name{}}:
We use $\tau=0.05$ as reported in the paper.

\textbf{Implementation of CLIP}:
We tune the $\tau$ within [0.1, 0.2, 0.5], 
and find that $\tau=0.1$ performs best, corresponding to the reported results.
The batch size of CLIP is 32*48 since we have to reduce the batch size to 
load the language model.
The effective batch size of CLIP is 32*48*5,
since we only have titles of the PV items in our dataset.
The text model is adapted from Chinese-BERT\cite{chinese_bert}
and is initialized with the pre-trained weights provided by the authors.

\textbf{Implementation of SimCLR}:
We tune the $\tau$ with in [0.1, 0.2, 0.5], 
and find that $\tau=0.2$ performs best, corresponding to the reported results.
The effective batch size of SimCLR is 64*48*10.
We implemented the same augmentation strategy as suggested by the SimCLR paper\cite{simclr}.

\textbf{Implementation of SimSiam}:
The effective batch size of SimSiam is 64*48*10.
The augmentation method is the same as SimCLR.

\section{Experiment on hinge loss}

From the perspective of metric learning\cite{deep_metric}, the current popular contrastive loss and the traditional hinge loss in metric learning serve similar purposes: they both aim to minimize the distance between similar embeddings. We agree that this is also a direction worth exploring.

Following our definition of similarity between PV item embeddings and reconstruction in \refeq{eq:sim}
\begin{equation}
    Sim(j_0, j_1) = \frac{{E_{pv}^{j_0}}^T R_{pv}^{j_1}}{||E_{pv}^{j_0}|| \; ||R_{pv}^{j_1}||}\label{eq:sim}
\end{equation}
Where $j_0$ enumerates over PV image embeddings, and $j_1$ enumerates over reconstructions.
We can define a hinge loss as
\begin{equation}
L_{hinge} = \max(0, Sim(j_0, j_1) - Sim(j_0, j_0) + m), j_1 \ne j_0 \label{eq:hinge_loss}
\end{equation}
Where $Sim(j_0, j_0)$ measures the cosine similarity between $E_{pv}^{j_0}$ and $R_{pv}^{j_0}$, and 
$Sim(j_0, j_1)$ measures the cosine similarity between $E_{pv}^{j_0}$ and $R_{pv}^{j_1}$ with $j_1 \ne j_0$, $m>0$ is a tunable hyper-parameter corresponding to margin. By definition, we want to make $Sim(j_0, j_0) > Sim(j_0, j_1)$. The hinge loss in \refeq{eq:hinge_loss} will equal to 0 of $Sim(j_0, j_0) > Sim(j_0, j_1) + m$, and equal to $Sim(j_0, j_1) - Sim(j_0, j_0) + m$ otherwise.

In triplet-based metric learning, sampling of positive and negative instances is required. In our scenario, positive instances have been constructed through user intention reconstruction, yet negative instances still need to be sampled. We randomly sample a mismatched reconstruction embedding from the batch as a negative instance to calculate the loss.

\begin{table}[h]
    \centering
    \caption{Train with hinge loss.}
    \begin{tabular}{lrrr}
        \toprule
               & \multicolumn{1}{l}{$\Delta \text{AUC}$ (women's clothing)} & \multicolumn{1}{l}{$\Delta \text{AUC}$} & \multicolumn{1}{l}{$\Delta \text{GAUC}$} \\
        \midrule
        Hinge loss & 0.15\%                                     & 0.02\%                  & 0.02\%                  \\
        \name  & \textbf{0.46\%}                            & \textbf{0.16\%}         & \textbf{0.19\%}          \\
        \bottomrule
    \end{tabular}%
    \label{tab:add_hinge}%
\end{table}%
We have tuned the implementation of this hinge loss, including parameters such as learning rate and margin parameter $m$. We keep other configurations the same as our method. \reftab{tab:add_hinge} presents the best results we have achieved.
The results obtained using triplet hinge loss are significantly worse than those obtained using contrastive loss, mainly for two reasons. Firstly, the performance of metric learning based on triplet hinge loss is highly sensitive to the sampling method\cite{hard_negative}, which may require complex sampling techniques to achieve better results. However, implementing complex sampling methods can increase training overhead, which is beyond the scope of our work. Secondly, contrastive loss allows adjusting the weight of negative examples by tuning the temperature $\tau$ parameter, effectively weighting hard negatives. This parameter tuning is simpler than implementing complex sampling methods and may lead to better results. Moreover, contrastive loss is more suitable for large-scale training.

\end{appendices}

\end{document}